\documentclass[letterpaper, 10 pt, conference]{ieeeconf}  %

\IEEEoverridecommandlockouts                              %

\usepackage{graphics} %
\usepackage{epsfig} %
\usepackage{mathptmx} %
\usepackage{times} %
\usepackage{amsmath} %
\usepackage{amssymb}  %
\usepackage{tabularx}  %
\usepackage{booktabs}
\usepackage{multirow}
\usepackage[ruled,noend,linesnumbered]{algorithm2e}
\usepackage[belowskip=0pt,aboveskip=5pt,font=small]{caption}
\usepackage[noadjust]{cite}
\usepackage[pagebackref=true,breaklinks=true, colorlinks,bookmarks=false,citecolor=red]{hyperref}
\usepackage{textcomp}
\newcommand{\til}{\raisebox{0.5ex}{\texttildelow}}

\title{\LARGE \bf
Success Weighted by Completion Time: \\ A Dynamics-Aware Evaluation Criteria for Embodied Navigation
}

\author{Naoki Yokoyama$^{1}$ \enspace\enspace Sehoon Ha$^{1,2}$ \enspace\enspace Dhruv Batra$^{1,3}$%
\thanks{$^{1}$NY, SH, DB are with the College of Electrical Engineering and College of Computing, Georgia Institute of Technology, Atlanta, GA 30318 USA. $^{2}$SH is with Robotics at Google. $^{3}$DB is with Facebook AI Research (FAIR). 
The Georgia Tech effort was supported in part by NSF, DARPA, ONR YIP, and ARO PECASE. 
(e-mail: nyokoyama@gatech.edu; sehoonha@gatech.edu; dbatra@gatech.edu)}%
}

\newcommand{\cmt}[1]{}

\long\def\ignorethis#1{}

\newcommand{\pctab}{\hspace{0.2in}}

\begin{document}

\maketitle
\thispagestyle{empty}
\pagestyle{empty}

\begin{abstract}
We present Success weighted by Completion Time (SCT), a new metric for evaluating navigation performance for mobile robots. Several related works on navigation have used Success weighted by Path Length (SPL) as the primary method of evaluating the path an agent makes to a goal location, but SPL is limited in its ability to properly evaluate agents with complex dynamics. In contrast, SCT explicitly takes the agent's dynamics model into consideration, and aims to accurately capture how well the agent has approximated the fastest navigation behavior afforded by its dynamics. While several embodied navigation works use point-turn dynamics, we focus on unicycle-cart dynamics for our agent, which better exemplifies the dynamics model of popular mobile robotics platforms (e.g., LoCoBot, TurtleBot, Fetch, etc.). We also present RRT*-Unicycle, an algorithm for unicycle dynamics that estimates the fastest collision-free path and completion time from a starting pose to a goal location in an environment containing obstacles. We experiment with deep reinforcement learning and reward shaping to train and compare the navigation performance of agents with different dynamics models. In evaluating these agents, we show that in contrast to SPL, SCT is able to capture the advantages in navigation speed a unicycle model has over a simpler point-turn model of dynamics. Lastly, we show that we can successfully deploy our trained models and algorithms outside of simulation in the real world. We embody our agents in a real robot to navigate an apartment, and show that they can generalize in a zero-shot manner. A video summary is available here: \url{https://youtu.be/QOQ56XVIYVE}
\end{abstract}
\section{INTRODUCTION}

Mobile robots have the potential to automate several tasks in the real world and alleviate the amount of work to be done by humans. But for mobile robots to be effective in the physical world, they must be able to effectively navigate previously unseen environments. Yet, without access to a map or floorplan, inferring an efficient path to a distant goal in a novel environment using only the robot's onboard sensors is a challenging task.

\begin{figure}[t]
    \centering
    \includegraphics[width=0.47\textwidth]{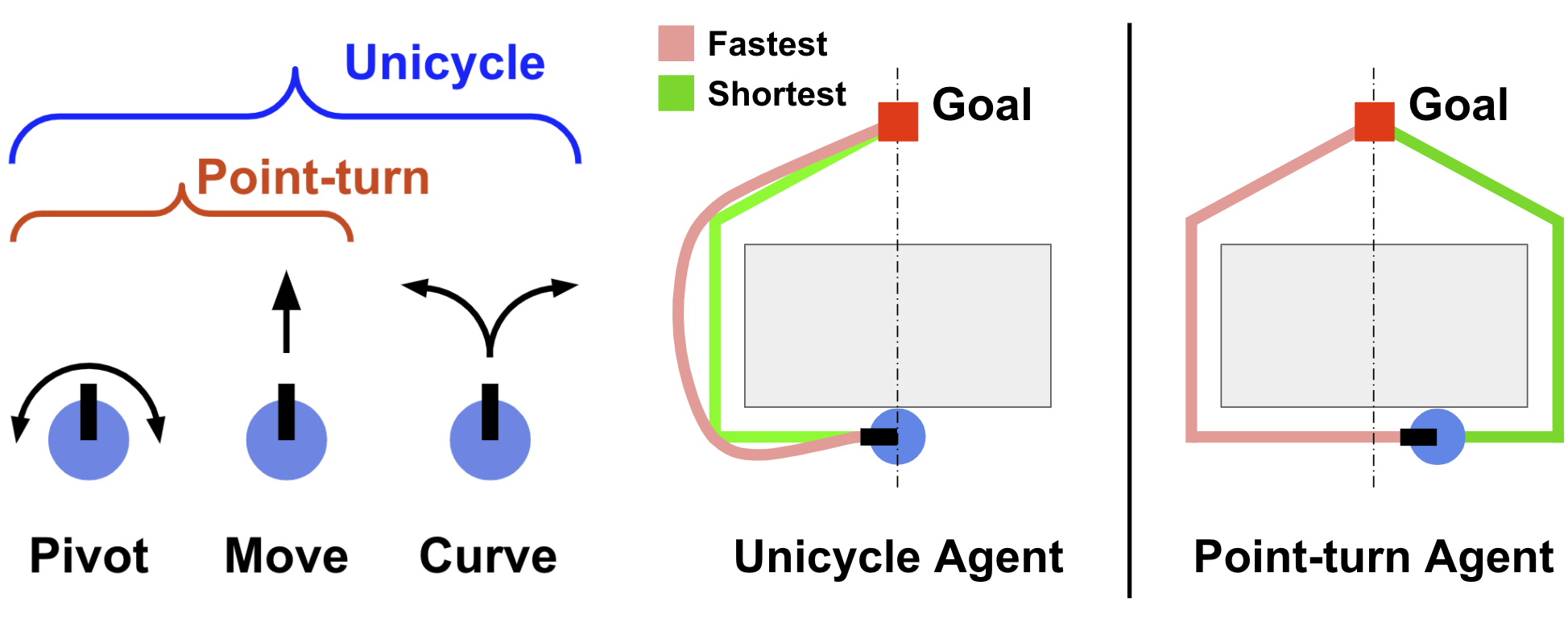}
    \includegraphics[width=0.475\textwidth]{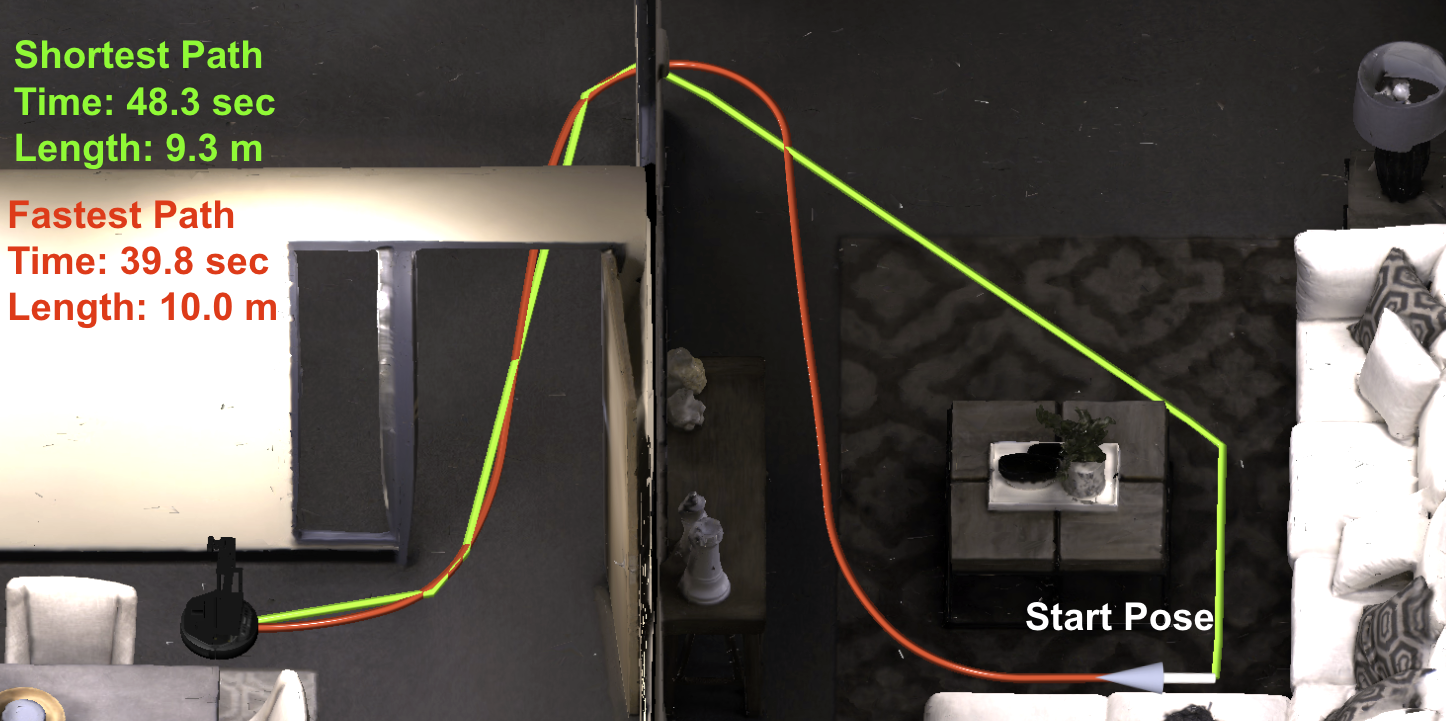}
    \caption{
The shortest path is not always the fastest. \textit{Top left:} In addition to pivoting and moving straight like a point-turn model, a unicycle-cart dynamics model can move forward and turn (curve) simultaneously. \textit{Top right:} A path faster than the shortest path may exist for both types of dynamics. The fastest path depends on the agent's maximum linear and angular velocities, and initial heading. \textit{Bottom}: Fastest path (calculated with RRT*-Unicycle) and shortest path visualized in Habitat, where our agents are trained.
    }
    \label{fig:unicycle_vs_pointturn}
\end{figure}

Several recent works have shown that deep reinforcement learning (DRL) can be used to teach virtual robots in simulators to accomplish tasks such as navigating to a goal point (\cite{ddppo, srcc, chaplot2020learning, habitat}), or searching for a distant object (\cite{batra2020objectnav, chaplot2020object}), even within previously unseen environments. However, these virtual robots are typically modeled as idealized cylinders, equipped with a limited high-level action space (move-forward 0.25m, pivot $\pm$10\textdegree). A subtle but important consequence of the forced-choice nature of such an action-space is that these idealized agents are assumed to have `point-turn' dynamics (illustrated in Figure \ref{fig:unicycle_vs_pointturn}), in which they must come to a full halt to turn. Deployment of such policies to real mobile platforms such as in \cite{srcc} and \cite{li2020unsupervised} results in motion that is often jerky and time-consuming, and can result in increased power consumption or localization errors that compound from abrupt motions, as shown in \cite{moon2010observation}. 

To address these challenges and navigate more efficiently, we propose to use a more complex model of dynamics, namely, a unicycle-cart model. An agent with unicycle dynamics can not only pivot in-place and move straight like a point-turn agent, but can also make smooth curves by moving with both linear and angular velocity. This allows it to execute smoother turns without halting to navigate around obstacles more quickly than a point-turn agent. These points are illustrated in Figure \ref{fig:unicycle_vs_pointturn}.

However, we find that navigation performance metrics such as Success weighted by Path Length (SPL) are unable to properly capture the benefits in speed brought by more complex dynamics models, and instead favors the jerky paths produced by point-turn behavior. Therefore, we propose a new metric, Success weighted by Completion Time (SCT), which addresses the shortcomings of navigation metrics such as SPL by explicitly taking into consideration the dynamics model of the agent. To calculate SCT, we formulate a method called RRT*-Unicycle to find the fastest path from a start to a goal for a unicycle model of dynamics, which can be extended to other models of dynamics. 

In this work, we use DRL to train agents with unicycle-cart dynamics to successfully navigate to the goal location significantly faster than those trained with point-turn dynamics, despite having the same maximum linear and angular velocities (i.e., not being able to move any faster). We show that SPL fails to capture the faster successful completion speeds of the unicycle agents, while SCT can. We also propose a decaying scheme for reward shaping, which we show results in agents learning faster navigation behavior than a fixed shaped reward function. Lastly, we deploy a unicycle agent on a real robot, and show that our model and reward scheme lead to trajectories that are 35\% faster than those of the point-turn agent with identical maximum linear and angular velocities.

\section{RELATED WORK}
\textbf{Visual Navigation with DRL.} Several works have also explored the use of DRL in training visual navigation policies using photorealistic 3D simulators of indoor environments, such as \cite{auto_navigator,aux_stefan,ddppo,chaplot2020object,li2020unsupervised, rosano2020embodied}. However, all of these works consider agents that choose from a small discrete set of actions, which only allow the agent to move forward or pivot in-place. Furthermore, many of these works evaluate their agents based on Success weighted by Path Length (SPL), a metric Anderson et. al recommended to adopt as the primary method of evaluating navigation performance in \cite{anderson}. In this work, we show why SPL is not optimal for evaluating agents with more complex dynamics, and propose a new metric to evaluate the navigation performance for such agents.

\textbf{Evaluation Metrics.} The metric proposed by \cite{chen2020semantic_audio_nav}, Success weighted by Number of Actions, closely resembles our proposed metric. Like our metric, it also penalizes redundant idle actions that do not necessarily increase the length of the agent's path (e.g., excessively turning in-place). However, it ultimately derives the baseline number of actions to compare against using the shortest path to the goal, while we focus on the fastest path conditioned on the agent's dynamics.
\section{PROBLEM FORMULATION}

\subsection{Task: Point Goal Navigation}
For this work, we consider the task of PointGoal Navigation (PointNav), as detailed in \cite{anderson}. In this task, the robot is initialized within a previously unseen environment, and must navigate to a goal point specified in relative coordinates. An \textit{episode} is defined by the environment to navigate, a starting location and heading, and a goal location. The agent must reach the goal point using only its onboard sensors as input. Our agents do not have access to a map; all maps and calculated paths shown in this work are for performance evaluation and illustrative purposes only.

\textbf{Success criteria.} Episodes are considered successful only if the agent is both within a certain radius of the goal point and invokes an action that terminates the episode. We choose the radius of LoCoBot \cite{locobot} (0.2m) as our success radius because it is the mobile platform that we use in our real-world experiments. Episodes are terminated and unsuccessful after 500 action steps if the agent has not terminated it yet.

\subsection{The limitations of SPL}
SPL scores the length of an agent's path based on how close it is to the length of the shortest path. Specifically,
\begin{align}
    \mbox{SPL}\ =\ S \frac{L}{\max{(P,L)}}
\end{align} 
where \textit{S} is 0 or 1 depending on whether the agent successfully completed the episode, \textit{P} is the length of the agent's path, and \textit{L} is the length of the shortest path from the start point to the goal point.

However, the shortest path to a goal is not necessarily the fastest. In many cases, the shortest path can actually be much slower or energy inefficient than an alternative path, as depicted in Figure \ref{fig:unicycle_vs_pointturn}. The fastest path depends highly on the agent's dynamics, as well as its initial heading at the start of the episode. Because SPL only evaluates the agent by comparing the length of its path to the shortest path length from the start to the goal, it incentivizes tight turns around obstacles without consideration for how inefficient following such a path may be for the agent's dynamics. This unfairly favors simpler point-turn behaviors of motion, and discourages agents to leverage their model of dynamics in any way that would deviate from that of a point-turn. Additionally, it does not penalize unnecessary or redundant actions that do not lengthen the agent's path, such as pausing or pivoting. This and other limitations of SPL are discussed in Batra et al \cite{batra2020objectnav}.

\subsection{Success weighted by Completion Time}
To overcome all of these limitations of SPL, we propose Success weighted by Completion Time (SCT), defined as:
\begin{align}
    \mbox{SCT}\ =\ S \frac{T}{\max{(C,T)}}
\end{align} 
where \textit{C} is the agent's completion time, and \textit{T} is the shortest possible amount of time it takes to reach the goal point from the start point while circumventing obstacles based on the agent's dynamics.

One alternative to SCT we considered was simply computing the agent's average completion time on successful episodes. However, computing the lower and upper bounds of the average completion time is not trivial, making it difficult to tell how well the robot \emph{could} have performed the task. It is also not comparable across datasets; for example, if one dataset contains navigation environments larger than another dataset's on average, its average successful completion time would likely be longer in comparison. SCT addresses these issues by measuring how close an agent's completion speed is to being optimal, using pre-computed fastest paths to compare the agent's paths against. 

\subsection{Calculating the fastest path time: RRT*-Unicycle}
To compute SCT, we need to compute the fastest possible completion time of an episode. Although a navigation mesh, an auxiliary data structure often provided by simulators representing the navigable areas of an environment, greatly simplifies the process of finding shortest paths using A*, they do not encode anything about a given agent's dynamics.  Therefore, it cannot be easily exploited to find the fastest path. For such a task, creating a new graph using RRT* \cite{rrtstar} that considers the agent's dynamics is often the preferred approach \cite{kino_rrt, noreen2016optimal}.

\begin{figure}[t]
    \centering
    
    \vskip5pt
    \includegraphics[width=0.48\textwidth]{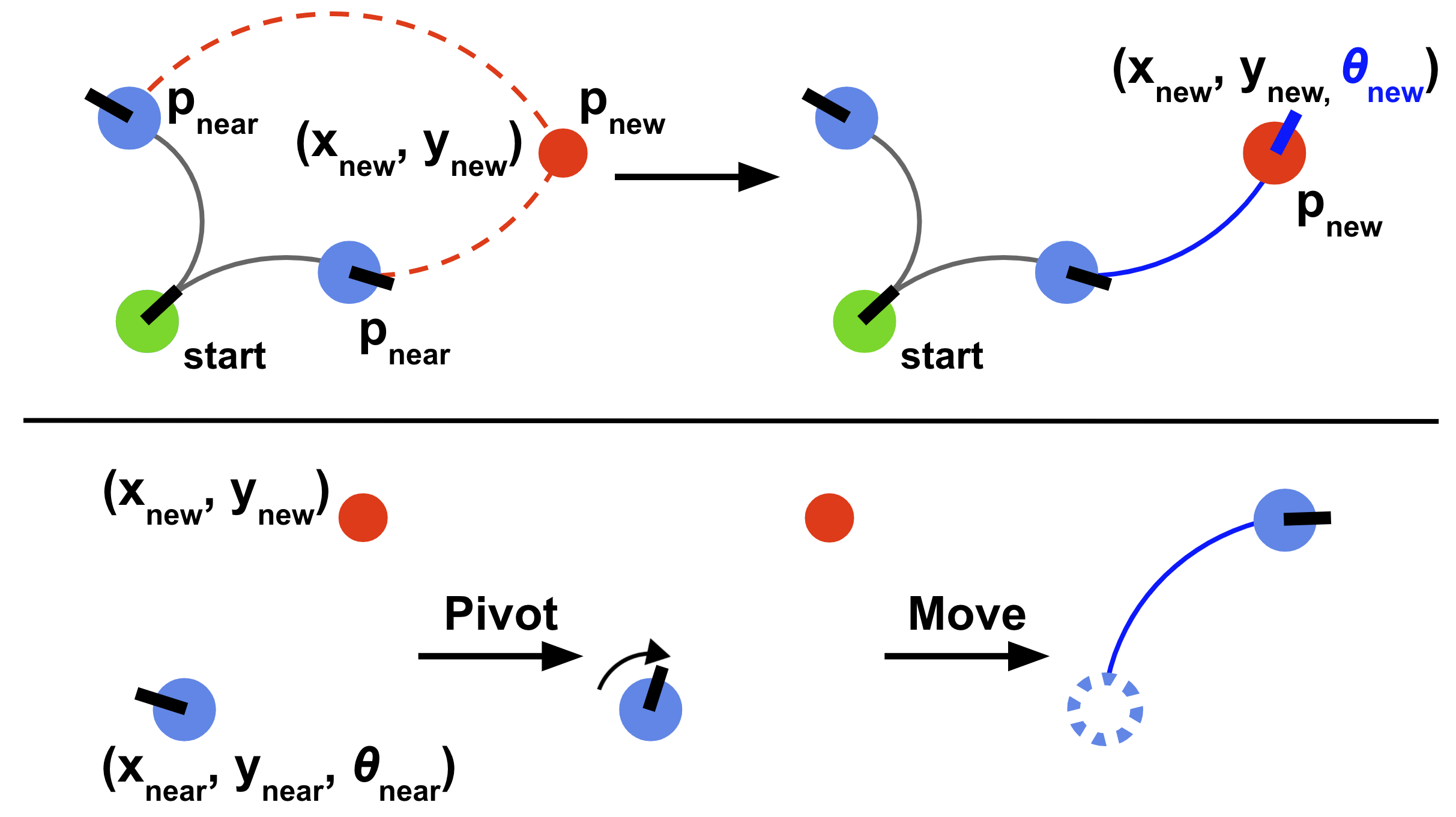}
    \caption{
\textit{Top:} The heading ($\theta$) of a new node is determined by the fastest path from the start that can be made from existing neighbors. \textit{Bottom:} We calculate the fastest path from a pose to a target location in free space by initially pivoting the agent towards the target with max angular velocity (if optimal), before moving towards the target location with max linear velocity and constant angular velocity.
    }
    \label{fig:fastest_path}
\end{figure}
We develop an adapted version of RRT*, which we call RRT*-Unicycle, to find the fastest path from the start to the goal of an episode that is conditioned on the robot's dynamics. Recall that RRT* is a sampling-based optimal motion-planning algorithm that, given a map, finds an obstacle-free path between two points that minimizes a cost, and is known to approach the optimal solution as the number of samples approaches infinity. The cost we aim to minimize is the time it takes the agent to travel from its starting pose to the goal point. RRT* works by creating a graph that traverses the free space of an environment, constructing and optimizing collision-free paths between sampled points until a full path from the start to the goal can be assembled. A full review of RRT* is beyond the scope of this document, and we refer the reader to \cite{rrtstar}. For the purpose of describing RRT*-Unicycle, we will be highlighting the changes made to the sampling, connecting, and rewiring stages of RRT*.

\textbf{Selecting a new node.} RRT*-Unicycle operates in 2D space, where each node $p$ in the constructed graph represents a pose of the agent $(x,\ y,\ \theta)$, defined by a location and heading.  When sampling a new node $p_{new}$, we initially leave $\theta_{new}$ uninitialized; only $(x_{new},\ y_{new})$ is fixed, which is randomly sampled from the free space. However, if $p_{new}$ is farther than a certain step distance \textit{r} from any existing node, we project it on to the circle with radius \textit{r} around the closest existing node. If $p_{new}$ is no longer in the free space (i.e., within an obstacle), we discard it and sample again. 

Similar to RRT*-Smart \cite{rrtstarsmart}, once a viable path from the start to the goal is found, we bias our sampling to the free space near this path to more rapidly approach the optimal solution. Additionally, when a viable path has not been found yet, we bias sampling around points on the shortest path (pre-computed using A* on the navigation mesh) to more efficiently explore the free space until a viable path has been found. We avoid explicitly adding the shortest path to the graph, as it is often unlikely to be the fastest path for a unicycle model.

\textbf{Connecting to an existing node.} Once $p_{new}$ is sampled, for each existing node $p_{near}$ within a certain radius of $p_{new}$, we calculate the fastest path from  $(x_{near},\ y_{near},\ \theta_{near})$ to $(x_{new},\ y_{new})$, as explained in the next subsection. The $p_{near}$ that yields the fastest path from the start to $p_{new}$ without colliding into any obstacles is chosen as the parent of $p_{new}$. To determine if a collision would occur, we uniformly sample points along a potential path and ensure that none intersect with an obstacle. When $p_{new}$ is finally added to the graph, $\theta_{new}$ is set as the heading of the agent when it arrived at location $(x_{new},\ y_{new})$ from the chosen $p_{near}$ via the fastest calculated path, as shown in the upper portion of Figure \ref{fig:fastest_path}.

\begin{figure}[t]
    \centering
    \vskip5pt
    \includegraphics[width=0.48\textwidth]{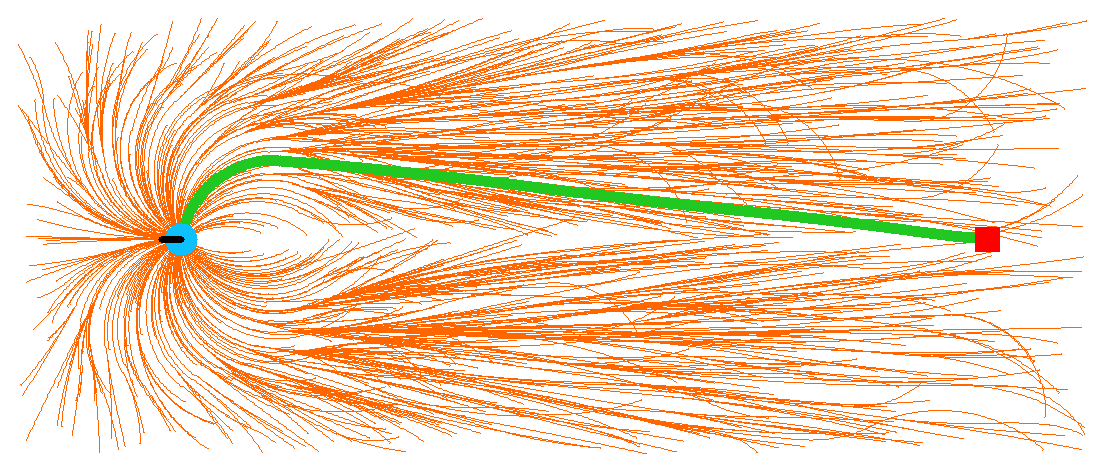}
    \caption{
Our method of finding the fastest path can link nearby nodes with circular arcs to approximate faster, more complex splines.
    }
    \label{fig:free_space}
\end{figure}

Leaving $\theta_{new}$ initially unfixed simplifies calculating the fastest path from a $p_{near}$ to $p_{new}$, as it is easier to calculate the fastest path from a source pose $(x_s,\ y_s,\ \theta_s)$ to a target \textit{location} $(x_t,\ y_t)$, than to a target \textit{pose} $(x_t,\ y_t,\ \theta_t)$. We find that allowing $\theta_{new}$ to instead be set by the fastest path from the neighboring node offering the shortest cost from the start point converges much faster towards the optimal path.

\textbf{Calculating the fastest path in free space.} To calculate the fastest path from a pose to a nearby location (assuming there are no obstacles), for simplicity, we assume that the fastest behavior is to follow an arc with maximum linear velocity and a constant angular velocity that would lead the agent directly to the target location from its current pose. An agent can perform an in-place pivot with maximum angular velocity towards the target location to begin the path only if it would decrease the total path completion time.  To determine how much pivoting yields the shortest path time, we used a brute-force method to generate a lookup table mapping the heading and distance relative to the target location to the optimal amount of pivoting needed before moving along the arc. This behavior is shown in the bottom of Figure \ref{fig:fastest_path}. Though the resulting circular arc is not always the fastest path, especially over longer distances, it is a good close approximation for nearby points. By joining enough nearby points together with this method, we can form the more complex splines between points in free space that represent faster paths than an arc, as shown in Figure \ref{fig:free_space}.

\textbf{RRT* Rewiring.} During the rewiring step, the nodes near $p_{new}$ disconnect from their parent node and adopt $p_{new}$ as its new parent if doing so decreases their cost to be reached from start node. Thus, we must calculate the cost from $p_{new}$ to a nearby $p_{near}$, where both nodes now have fixed $\theta$ values. In other words, the fastest path time must be calculated from a pose to a pose, rather than from a pose to a location. To calculate this cost, we simply calculate the fastest path and its completion time from the source pose to the target poses's location, as described previously, and then add the amount of time necessary to pivot in-place until $\theta_{near}$ is reached as fast as possible. Though this is an overestimation of the true fastest path between two poses, we find that this method is still able to lead to a near-optimal path with enough samples.

\textbf{Extracting the best path.} To get the best path to the goal from the tree, for each node $p_{near}$ within a certain radius of $(x_{goal},\ y_{goal})$, we compute the cumulative fastest path completion time from the start node $p_{start}$ to $p_{near}$, add the completion time of the fastest path from $p_{near}$ to $(x_{goal},\ y_{goal})$, and select the $p_{near}$ that minimizes this sum. The path from $p_{start}$ to the selected $p_{near}$ joined with the fastest path from $p_{near}$ to $(x_{goal},\ y_{goal})$ represents the fastest path from the start pose to the goal location found by the tree. 

\begin{figure}[t]
    \centering
    \vskip5pt
    \includegraphics[width=0.48\textwidth]{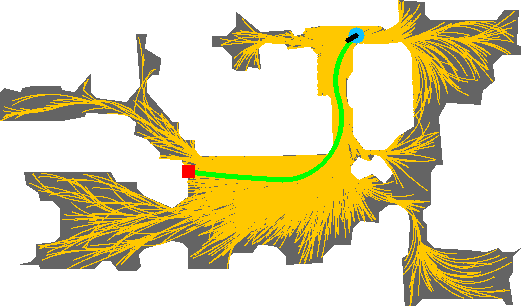}
    \caption{
A tree created with RRT*-Unicycle for calculating the fastest path. Fastest found paths to all sampled points shown in yellow, and fastest found path to the goal in green. These paths consider the maximum linear and angular velocities of our unicycle-cart agents, as well as its initial heading at the start of the episode. Neither the maps nor the paths are provided to the agents during training or testing; they are for performance evaluation purposes only.
    }
    \label{fig:rrt_visualization}
\end{figure}

\textbf{Summary.} The full procedure is summarized in Algorithm \ref{alg:rrt_star}. An example of a constructed tree where the fastest paths to all sampled nodes are visualized is shown in Figure \ref{fig:rrt_visualization}, where the fastest path to the goal point is highlighted. Notice that the fastest path is not the shortest.

Though we focus RRT*-Unicycle on a unicycle-cart model of dynamics, which reflects the type of motion widely used by real indoor mobile robots, it can easily be extended to other dynamics models as well. This can be done by changing the way in which the fastest path from a pose to a target location under an obstacle-free setting is calculated, which defines the cost of each edge in the graph.

\SetAlFnt{\fontsize{9.5}{0}} 
\begin{algorithm}
\caption{RRT*-Unicycle}
\label{alg:rrt_star}
\SetAlgoNoLine%
\newcommand\mycommfont[1]{\footnotesize\ttfamily\textcolor{blue}{#1}}
\SetCommentSty{mycommfont}
\SetArgSty{textnormal}
E $\leftarrow$ 0.2 \tcp{Exploration probability}
$T\leftarrow$InitializeTree$(x_{start}, y_{start}, \theta_{start}, x_{goal}, y_{goal}$)\\
\For{$i$ = 0, ..., N}{
    $e \leftarrow \text{random}()$\\
    \uIf{$e<E\ \text{\textbf{or\ not}}\ T.\text{pathToGoalFound}()$} {
        \uIf(\tcp*[h]{Sample free space}){$e<E$} {
            $x_{new},\ y_{new} \leftarrow $ sampleNavigableXY()\\
        } 
        \Else(\tcp*[h]{Sample near shortest path}){ 
            $x_{new},\ y_{new} \leftarrow $ sampleNearShortestPath()\\
        }
        $x_{new},\ y_{new} \leftarrow T$.project($x_{new},\ y_{new}$) \label{lst:line:steer}
    } 
    \Else (\tcp*[h]{Sample near current fastest path}){
        $x_{new},\ y_{new} \leftarrow T.$sampleNearFastestPath()
    }
    $p_{near}^{1\cdots N} \leftarrow T$.nearbyNodes($x_{new},\ y_{new}$)\\
    $p_{near},\ \theta_{new}\leftarrow$pickParent($x_{new},\ y_{new},\ p_{near}^{1\cdots N})$\\
    $p_{new}\leftarrow T.$InitializeNode($x_{new},\ y_{new},\ \theta_{new}$)\\
    $T$.addEdge($p_{near},\ p_{new}$)\\
    $T$.rewire($p_{new}$)\\ 
}
\Return $T$.fastestPath()\\
\end{algorithm}
\DecMargin{1em}

\section{Experimental Setup}

\subsection{Simulator and dataset}\label{dataset}

We train our agents within Habitat \cite{habitat}, a high-performance physics-enabled 3D simulator that provides the virtual robot with first-person photorealistic visual data. We use the same simulator settings as the Habitat 2019 challenge \cite{habitat}. For training and evaluation, we use the Gibson-4+ dataset \cite{habitat}, comprised of 86 high quality 3D scans and meshes curated from the full Gibson dataset \cite{Xia_2018_CVPR}. The Gibson dataset includes scans of various indoor environments, such as apartments, multi-level homes, offices, houses, hospitals, and gyms. The Gibson-4+ dataset only contains scenes that have been manually rated 4 or above on a quality scale from 0 to 5, and are free of significant reconstruction artifacts such as holes in texture or cracks in floor surfaces. In total, the training split of Gibson-4+ contains 3.6M episodes distributed across the 72 scenes, while the validation split contains 994 episodes across 14 scenes. We divide the validation split into a val$_1$ and val$_2$ set by randomly selecting 7 out of the 14 scenes for each set (497 episodes in each set). After training each agent on the train split, we choose the policy weights for each agent based on their best SCT performance on val$_1$. Each agent then uses these weights to perform the episodes of val$_2$, and performance on val$_2$ is used to report evaluation metrics.

\subsection{Action space}
We use discrete action spaces for training both point-turn and unicycle agents. At each step, the policy is polled and outputs an action representing a pair of linear and angular velocities. The simulator has the agent assume its output velocities (unless it collides) until it is polled again. The policy is polled at 1 Hz. The agent terminates the episode by outputting zero linear and angular velocity. We constrain the linear velocity to be non-negative, as we have found that allowing the agent to move backwards hurts performance; the resulting behavior often moves backwards excessively and does not use information from the agent's depth camera effectively, as the agent's camera is forward-facing.
\begin{figure}[t]
    \centering
    \vskip5pt
    \includegraphics[width=0.45\textwidth]{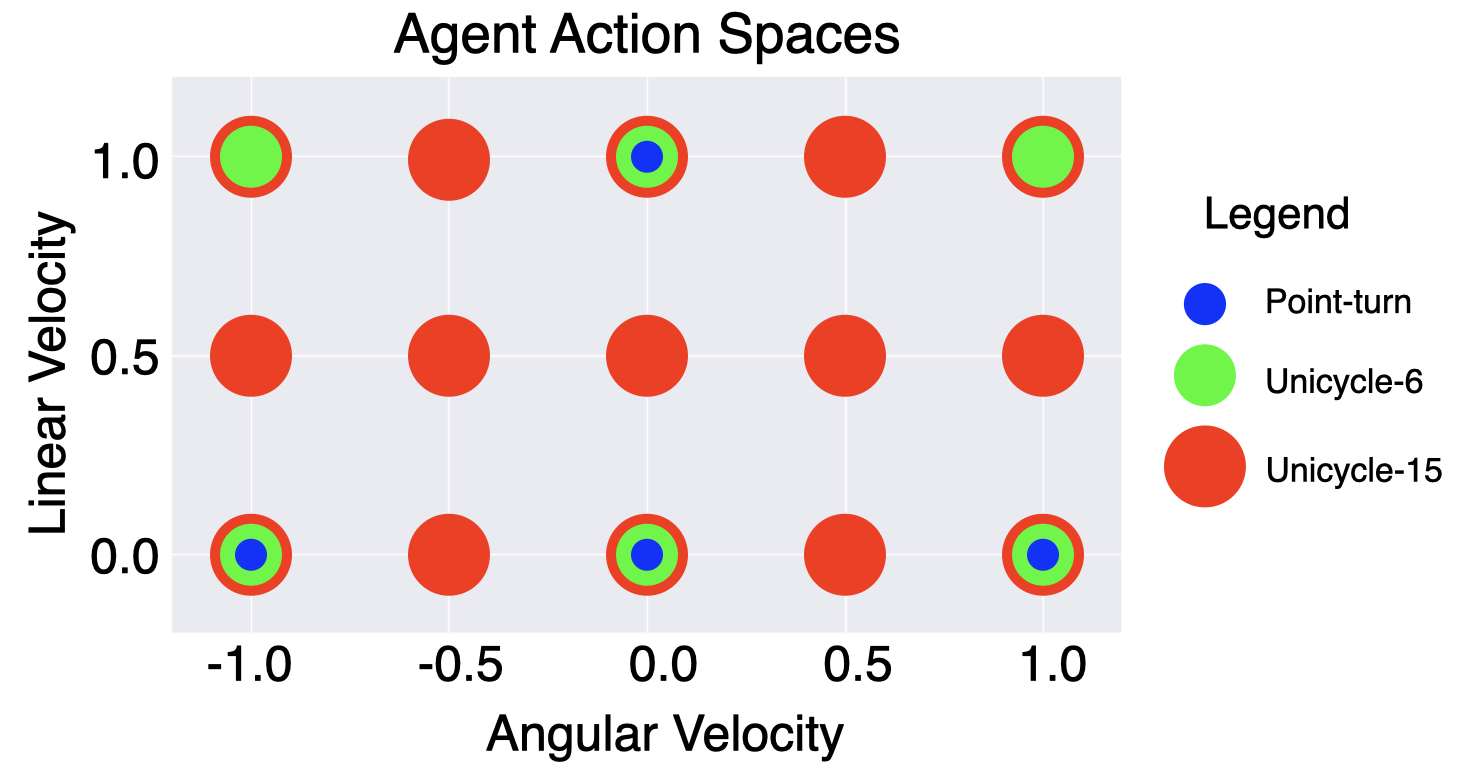}
    \caption{
Discrete action space used by each agent. Actions are scaled by the agent's maximum linear and angular velocities.
    }
    \label{fig:action_space}
\end{figure}

As shown in Figure \ref{fig:action_space}, we use discrete action spaces with different amounts of actions to model different dynamics:
\begin{itemize}
  \item Point-turn (2 lin. vel. + 3 ang. vel. - 1 overlap = 4 actions)
  \item Unicycle-6 (2 lin. vel. $\times$ 3 ang. vel. = 6 actions)
  \item Unicycle-15 (3 lin. vel. $\times$ 5 ang. vel. = 15 actions)
\end{itemize}
We set the maximum linear velocity to 0.25m/s and the maximum angular velocity to 10\textdegree/s. For the actions of the point-turn agent, this translates to 0.25m or 10\textdegree\ of displacement per step, which is used by other works such as \cite{auto_navigator, ddppo, li2020unsupervised, rosano2020embodied}.

\subsection{Observation space}\label{sensors}

The robot is equipped with an egocentric depth camera and an egomotion sensor. The robot's depth camera provides single channel 256$\times$256 images where each pixel represents a depth of up to 10m. The egomotion sensor provides the agent with its current displacement and relative heading from its initial start position. This is used to calculate the relative heading and distance to the goal from the goal coordinates, which is then given to the robot's policy.  

\subsection{Policy model}
\begin{figure}[t]
    \centering
    \vskip5pt
    \includegraphics[width=0.48\textwidth]{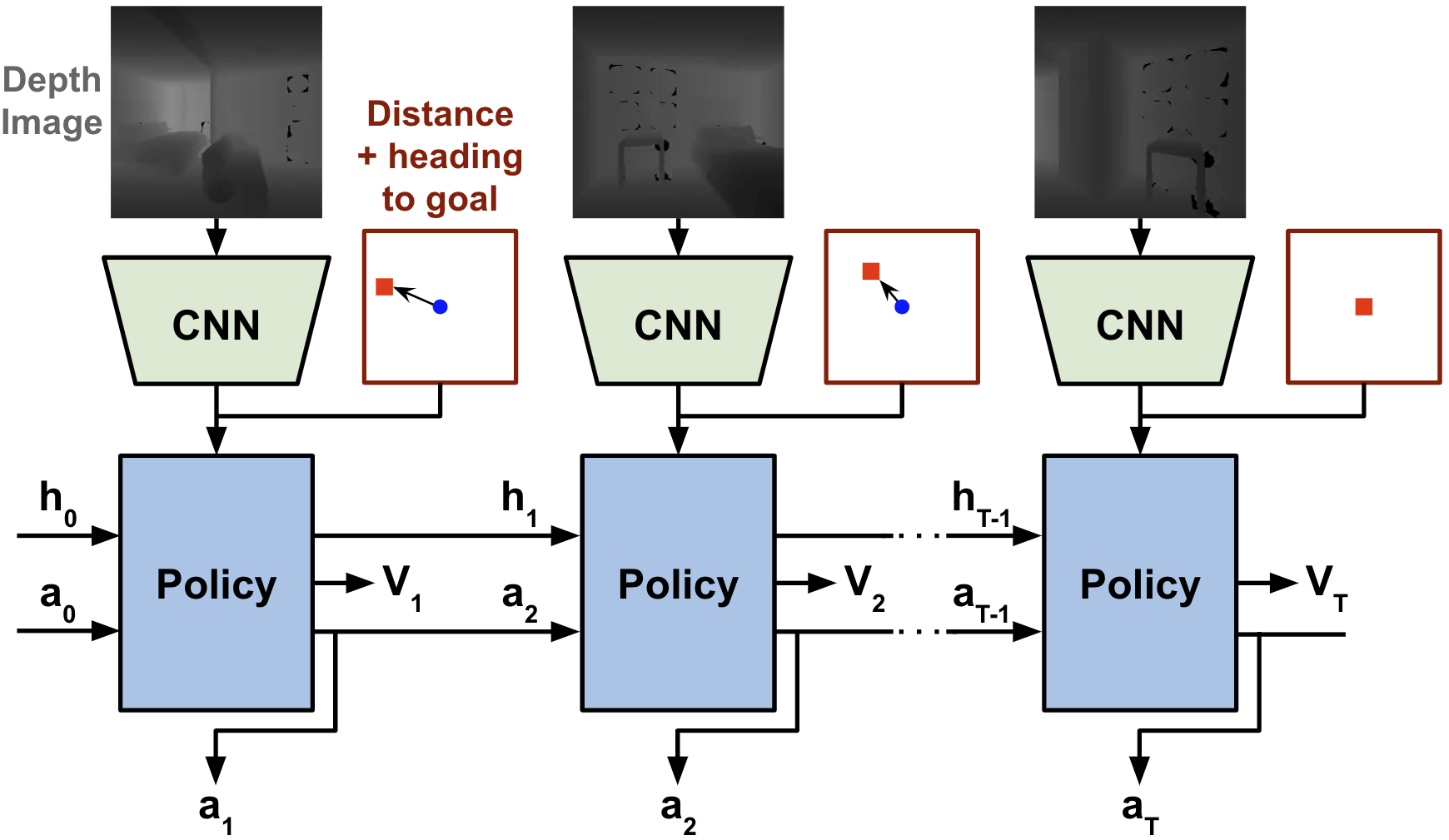}
    \caption{
Network architecture for our agents, comprised of a convolutional encoder and a recurrent policy. The agent receives an egocentric depth image, its current distance and heading relative to the goal point, and its previous action at each time step. It outputs an action and an estimate of the value function.
    }
    \label{fig:model}
\end{figure}

We use the neural net architecture described in \cite{ddppo} to model the policy of each of our agents. It has two main components; a visual encoder and a policy. The visual encoder is a convolutional neural network based on ResNet-50 \cite{resnet}, and takes the depth image as input. The policy is comprised of a 2-layer LSTM recurrent neural network with a 512-dimensional hidden state, taking the visual encoder's features, the relative distance and heading to the goal, the previous action, and the previous hidden state as input (Figure \ref{fig:model}). The goal distance and heading are represented as $[r,\ cos(\theta),\  sin(\theta)]$ to avoid the discontinuity at 180\textdegree, and the previous action is represented as a one-hot encoded vector, whose length depends on the agent's action space. The final layer of the policy has either 4, 6, or 15 outputs (depending on the agent's action space), which parameterize a categorical action distribution from which to sample from. The policy also has a critic head that outputs an estimate of the state's value, which is used for reinforcement learning as described in \cite{ddppo}. 

\begin{figure*}[t]
    \centering
    \vskip5pt
    \includegraphics[width=0.99\textwidth]{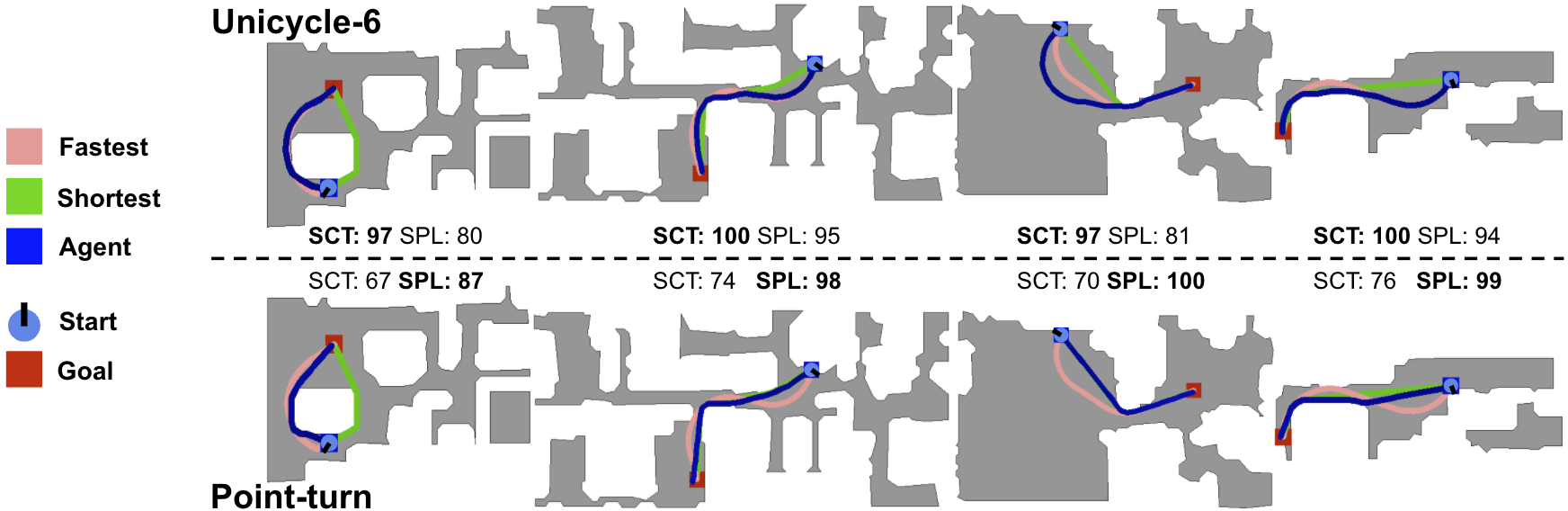}
    \caption{
The unicycle agents learn to leverage their dynamics to complete episodes faster than point-turn agents (higher SCT) by using smoother turns, which leads to longer paths and consequently lower SPL. Visualization is of performance on example episodes from val$_2$, trained with the decaying reward.
    }
    \label{fig:habitat}
\end{figure*}

\subsection{Training and reward}

We use Decentralized Distributed Proximal Policy Optimization by Wijmans et al. \cite{ddppo} to train our agents using deep reinforcement learning, and use the same training hyperparameters. We train 8 copies of the agent running in parallel per GPU, using 8 GPUs for a total of 64 parallel workers, for 150M cumulative steps (\til240 GPU hours, or \til30 hours wall-clock).

We consider two different reward schemes. The first reward scheme is based on the one used by Wijmans et al., which has shown to yield very good performance ($>$96\% success, $>$0.91 SPL) using a point-turn agent. However, to make learning easier, it uses a shaped step reward that explicitly encourages the agent to follow the shortest path. Under such a reward scheme, the agent may not learn to take a path that deviates from the shortest path, even if it would allow the agent to reach the goal faster.  Because we aim to evaluate the agent based on how well it approximates the fastest completion behavior with SCT, we thus consider a second reward scheme that ``un-shapes'' this step reward, by gradually reducing its shaped portion to 0 as training progresses.

\textbf{Shaped reward scheme.} In the reward scheme used by Wijmans et al., the agent receives a shaped reward $r(a_t,s_t)$ at each time step based on geodesic distance, which is defined as the length of the shortest obstacle-free path between two points. We compute geodesic distance using A* on a navigation mesh provided by the simulator. When the agent reduces the geodesic distance from its position to the goal point, the change in geodesic is negative; thus, we multiply by -1 to produce a positive shaped reward if the agent reduces its shortest route to the goal. A slack penalty (-0.01) is added to discourage the robot from taking unnecessary actions. Lastly, a terminal reward $r_T$ is given at the end of the episode if the goal has been reached. Though Wijmans et al. scale the terminal reward by the agent's SPL for the episode, we avoid this to refrain from explicitly penalizing agents for deviating from the shortest path. The full reward scheme is defined as:
\begin{align}
    r(a_t,s_t)\ =\ -\Delta_{geo\_dist}-0.01,\ \  
    r_T\ =\ 2.5S
\end{align}
where S is 0 or 1 depending on whether the agent successfully completed the episode.

\textbf{Decaying reward scheme.} As previously mentioned, our aim with the second reward scheme is to prioritize successful completion speed. We keep the terminal reward the same, but slowly phase the shaped portion of the step reward to 0 as training progresses:
\begin{align}
    r_{decay}(a_t,s_t)\ =\ -\beta\Delta_{geo\_dist}-0.01,\ \  
    r_T\ =\ 2.5S
\end{align}
where $\beta$ starts with a value of 1, and linearly decays to 0 as the amount of elapsed training steps reaches a certain amount (80 million steps was chosen for our experiments). The goal of this reward scheme is to eventually provide the agent with only a negative constant step reward and a large positive constant terminal reward; the latter encourages the agent to complete the episode successfully, while the former encourages it to do it as quickly as possible. We do not immediately remove the geodesic portion of the reward at the start of training because the reward may be too sparse to learn optimal behaviors. Specifically, an agent early in training that has likely not learned to consistently reach the goal yet would be encouraged to simply terminate the episode immediately, such that it can avoid accruing the negative step reward. The geodesic term provides the positive reward signal necessary, even during the early stages of training, to lead the agent to the goal and teach it about the existence of a large positive terminal reward. Once the agent has learned that continuing the episode can lead to a large reward, we find that it continues to do so even as the geodesic term is completely eliminated.

\section{Results}
We aim to address the following questions:
\begin{enumerate}
  \item How effective are SCT and SPL for capturing the navigation efficiency of agents with a dynamics model more complex than a point-turn model?
  \item How does un-shaping the reward function affect SCT and other metrics of performance across different dynamics?
  \item Do our methods and results generalize to reality?
\end{enumerate}
\subsection{Evaluation details}
After training, we set the agent's behavior to be deterministic for both the val$_1$ and val$_2$ sets, meaning that the action with the highest likelihood in the output action distribution is always chosen. The agent does not receive any rewards, relying entirely on its depth camera and egomotion sensor as input to reach the goal. As detailed in Section \ref{dataset}, all reported evaluation metrics are based on performance on val$_2$. All reported SCT, SPL, and success rate values are multiplied by 100 for readability. To calculate SCT, we use RRT*-Unicycle to approximate the fastest possible path time for a unicycle-cart model of dynamics constrained to a maximum linear velocity of 0.25m/s and an angular velocity of 10\textdegree/s. Although the fastest completion times of SCT are computed based on unicycle dynamics, calculating SCT for the point-turn agent in the same way as the unicycle agents allows us to properly evaluate how much faster or slower it is successfully completing the test episodes in comparison.

\subsection{Path Length vs. Completion Time}
The results of our experiments within simulation are shown in Table \ref{tab:results_table}. Average SCT and SPL values are highly correlated with success rate, as unsuccessful episodes will yield 0 SCT and 0 SPL. This complicates the comparison of agents with \textit{varying success rates} in terms of average path length and completion time. We thus remove success rate as a variable by re-calculating average SCT and SPL using only the episodes that \textit{all} 6 agent variants (3 agents $\times$ 2 reward schemes) succeeded in (totaling to 87\% of val$_2$), dubbing them SCT$^\cap$ and SPL$^\cap$. These values show that for successful episodes, the unicycle agents completed episodes much faster than the point-turn agent ($>$88 SCT$^\cap$ vs. \til 72 SCT$^\cap$), but took paths that were not as short (\til 90 SPL$^\cap$ vs. \til 95 SPL$^\cap$).
\setlength{\tabcolsep}{3pt}
\begin{table}[h]
\centering
\vskip5pt
\caption{
Performance comparison between Point-turn, Unicycle-6, and Unicycle-15 agents using the shaped and decaying reward schemes. Mean and 95\% confidence interval are reported.
}
\label{tab:results_table}
\resizebox{\columnwidth}{!}{
    \begin{tabular}{ccccccc}
        \toprule
        \textbf{Agent}  & \textbf{Decay} & \textbf{SCT $\uparrow$} & \textbf{SPL $\uparrow$} & \textbf{Succ. $\uparrow$} & \textbf{SCT$^{\cap} \uparrow$} & \textbf{SPL$^{\cap} \uparrow$} \\
        \hline
        \multirow{2}{*}{PT} 
        & - & 65.82\tiny{$\pm$}1.95 & \textbf{88.41\tiny{$\pm$}2.18} & \textbf{94.16\tiny{$\pm$}2.06} & 71.26\tiny{$\pm$}1.33 & 95.11\tiny{$\pm$}0.92 \\
        & \checkmark & \textbf{65.99\tiny{$\pm$}2.07} & 87.91\tiny{$\pm$}2.28 & 93.96\tiny{$\pm$}2.09 & \textbf{71.75\tiny{$\pm$}1.45} & \textbf{94.90\tiny{$\pm$}1.08} \\
        \hline
        \multirow{2}{*}{U-6} 
        & - & 82.15\tiny{$\pm$}2.56 & 82.83\tiny{$\pm$}2.42 & 93.16\tiny{$\pm$}2.22 & 90.41\tiny{$\pm$}1.59 & 90.70\tiny{$\pm$}1.35 \\
        & \checkmark & \textbf{88.92\tiny{$\pm$}1.84} & \textbf{88.07\tiny{$\pm$}1.61} & \textbf{98.79\tiny{$\pm$}0.96} & \textbf{93.66\tiny{$\pm$}1.19} & \textbf{91.64\tiny{$\pm$}1.02} \\
        \hline
        \multirow{2}{*}{U-15} 
        & - & 81.43\tiny{$\pm$}2.53 & 83.31\tiny{$\pm$}2.26 & 94.37\tiny{$\pm$}2.03 & 88.37\tiny{$\pm$}1.82 & 89.69\tiny{$\pm$}1.41 \\
        & \checkmark & \textbf{86.01\tiny{$\pm$}2.10} & \textbf{85.37\tiny{$\pm$}1.87} & \textbf{95.77\tiny{$\pm$}1.77} & \textbf{91.09\tiny{$\pm$}1.56} & \textbf{89.66\tiny{$\pm$}1.34} \\
        \bottomrule
    \end{tabular}
    }
\end{table}

This finding supports our argument that SPL may not be a suitable metric for robots with more realistic dynamics. For a model of dynamics such as the unicycle-cart, it is unclear what value of SPL is `optimal'. On one hand, the maximum value of 100 SPL would restrict a unicycle agent's behavior to only move with its point-turn actions. On the other, targeting a lower SPL value that better corresponds to a learned behavior that more efficiently reaches the goal by leveraging non-point-turn dynamics is difficult, as it would be hard to determine an appropriate target SPL. Since SCT uses fastest completion times conditioned explicitly on the agent's dynamics as a baseline, it is a better metric for directly measuring how well the agent's learned behavior leverages the extents of its dynamics for navigation.

\subsection{Reward scheme comparison}
Table \ref{tab:results_table} shows that relative to the shaped reward scheme, our decaying reward scheme significantly increased the average SCT of the unicycle agents (82.2$\rightarrow$88.9 SCT for Unicycle-6, 81.4$\rightarrow$86.0 SCT for Unicycle-15), even with increases in success rate (93.2\%$\rightarrow$98.8\% for Unicycle-6, 94.4\%$\rightarrow$95.8\% for Unicycle-15) removed as a variable (90.4$\rightarrow$93.7 SCT$^\cap$ for Unicycle-6, 88.4$\rightarrow$91.1 SCT$^\cap$ for Unicycle-15). On the other hand, the performance of the point-turn agent did not change much between reward schemes ($<$0.5 change for SCT/SPL/SCT$^\cap$/SPL$^\cap$, 0.2\% change in success rate). We believe that this is because the original reward scheme encourages the agent to make shorter paths to the goal, while the decaying reward scheme encourages faster ones; these paths are not much different for a point-turn agent, so the learned behavior would not change much. Our results show that by gradually removing explicit encouragement of following the shortest path, the unicycle agents learn to better leverage the capabilities of their dynamics to complete the episodes faster and succeed more frequently.

\begin{figure}[t]
    \centering
    \vskip5pt
    \includegraphics[width=0.485\textwidth]{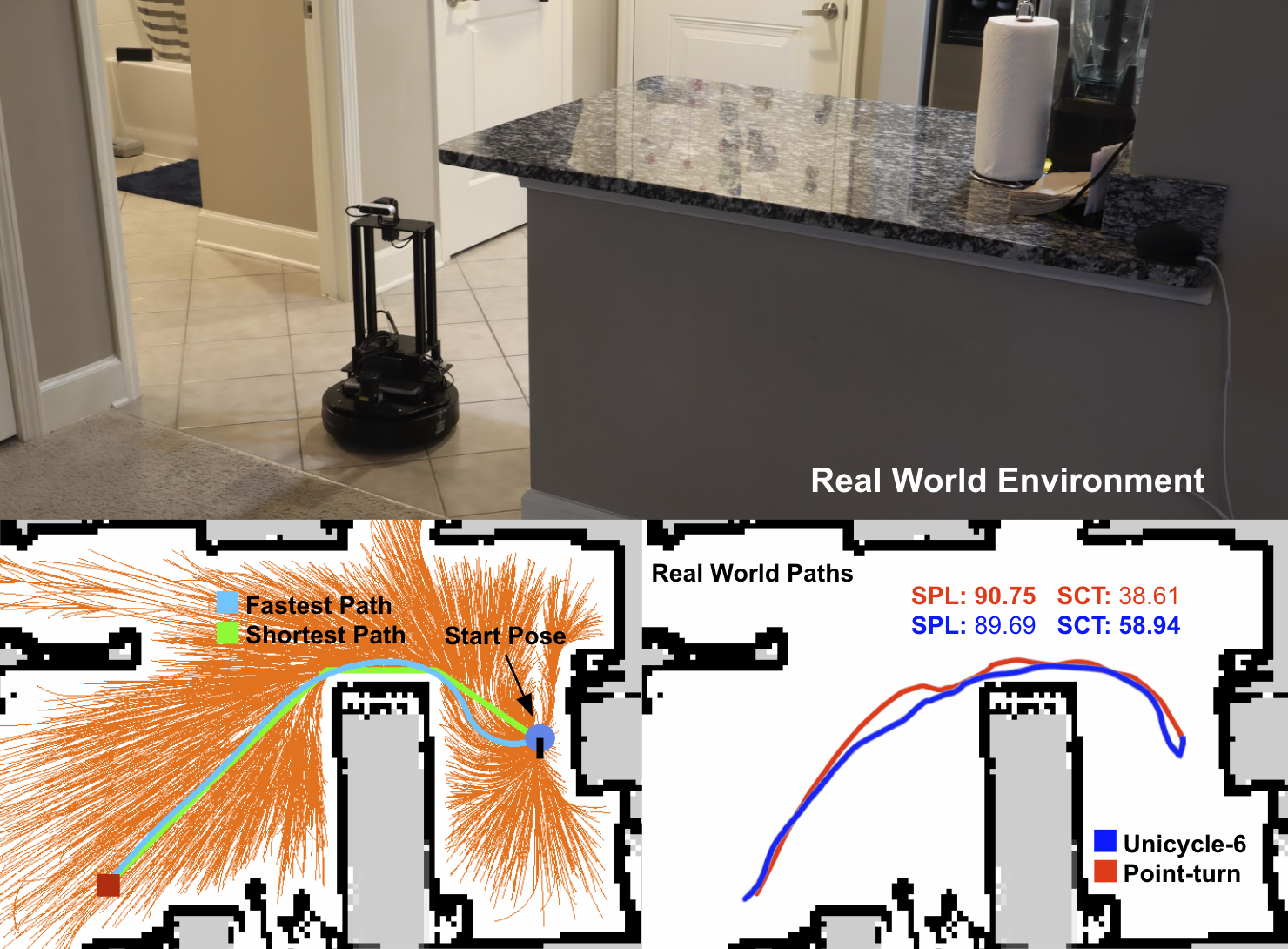}
    \caption{
\textit{Top:} The LoCoBot mobile robot and apartment used in real-world experiments. \textit{Bottom left:} RRT*-Unicycle and RRT*-Vanilla used on a 2D LiDAR scan to find the fastest completion time and shortest path length, respectively. \textit{Bottom right:} Example runs from the two agents on Ep. ID \#1 shown on the right.
    }
    \label{fig:reality_map}
\end{figure}

\subsection{Real-world experiments}

\begin{table}[ht]
\centering
\caption{Results from reality experiments. Each episode is the average of three independent trials. The Unicycle-6 agent achieves a higher SCT and faster completion time, while the Point-turn agent achieves a higher SPL and shorter distance travelled.}
\label{tab:results_reality}
\resizebox{\columnwidth}{!}{
    \begin{tabular}{cccccc}
        \toprule
        \textbf{Ep.} & \textbf{Agent} & \textbf{SCT $\uparrow$} & \textbf{SPL $\uparrow$} & \textbf{Time} & 
        \textbf{Agent}\\
        \textbf{ID} & & & & \textbf{(sec.) $\downarrow$}& \textbf{Dist. (m) $\downarrow$} \\
        \hline
        \multirow{2}{*}{1} & PT & 36.80\tiny{$\pm$}2.13 &  \textbf{88.37\tiny{$\pm$}2.82} & 61.00\tiny{$\pm$}3.61 & 
        \textbf{5.59\tiny{$\pm$}0.18}\\
        & U-6 & \textbf{54.01\tiny{$\pm$}4.62} & 85.46\tiny{$\pm$}3.99 & \textbf{41.67\tiny{$\pm$}3.51} &
        5.78\tiny{$\pm$}0.27\\
        \hline
        \multirow{2}{*}{2} & PT & 37.99\tiny{$\pm$}5.60 & \textbf{93.79\tiny{$\pm$}0.81} & 
        86.00\tiny{$\pm$}13.86 &
        \textbf{8.03\tiny{$\pm$}0.07}\\
        & U-6 & \textbf{59.96\tiny{$\pm$}6.05} & 92.87\tiny{$\pm$}3.49 & 
        \textbf{54.00\tiny{$\pm$}5.57} &
        8.11\tiny{$\pm$}0.31\\
        \hline
        \multirow{2}{*}{Avg.} & PT & 37.39\tiny{$\pm$}3.84 & \textbf{91.08\tiny{$\pm$}3.50} & 
        73.50\tiny{$\pm$}16.42 &
        \textbf{6.81\tiny{$\pm$}1.34}\\
        & U-6 & \textbf{56.98\tiny{$\pm$}5.81} & 89.16\tiny{$\pm$}5.27 & 
        \textbf{47.83\tiny{$\pm$}7.94} &
        6.95\tiny{$\pm$}1.30\\
        \bottomrule
    \end{tabular}
    }
\end{table}
We conducted real-world experiments in an indoor apartment to test whether point-turn and unicycle policies trained in simulation using our decaying reward scheme can generalize to reality. We used the LoCoBot mobile robot platform \cite{locobot}, equipped with a Hokuyo UTM-30LX LiDAR. The raw LiDAR data is \textit{not} processed by the robot's policy, and is used only to provide local egomotion estimates (relative heading and location from start position). The robot is also equipped with an Intel D435 depth camera to provide the policy with depth imagery. To calculate the completion time of the fastest path, we use RRT*-Unicycle on the floorplan generated by the Hector SLAM algorithm \cite{hector_slam} using the LiDAR. Neither the calculated path nor the floorplan is provided to the agent, and is used for performance evaluation purposes only. Because a navigation mesh of the apartment is not available, we simply use a vanilla version of RRT* to find the shortest path and its length for SPL calculation. These paths are visualized in Figure \ref{fig:reality_map}.

We tested two navigation agents, Point-turn and Unicycle-6. We increase their maximum angular speed during training and testing to 30\textdegree/s, as we have found that commanding 10\textdegree/s is too slow, often resulting in little to no movement on the real LoCoBot. This translates to 30\textdegree\  per step (as the policy is polled at 1 Hz), which has shown to work for visual navigation using the LoCoBot in \cite{srcc}. We do not test with the Unicycle-15 agent, as its action space would contain angular speeds of 15\textdegree/s, which we also found to be too slow. Following the recommendations from Kadian et al. \cite{srcc}, we train these agents with Habitat's sliding setting set to `off' in the simulator settings to help minimize collisions with obstacles. Sliding is a setting which allows the agents to `slide' against walls and obstacles upon collisions, a behavior often used in video games and other physics-based simulators. Kadian et al. showed that agents trained with this setting turned on often abuse these collision dynamics in simulation to reach the goal, and collide much more often when deployed in the real world. When sliding is off, the agent does not move upon collision. Otherwise, training and checkpoint selection was done in the same manner as our 10\textdegree/s agent experiments reported in the previous subsections.

Each agent was tested on two different episodes, each with 3 independent trials, for a total of 12 real-world runs. A top-down view of one episode is shown in Figure \ref{fig:reality_map}. Note that our agents were trained solely in simulation, and were not fine-tuned to the evaluation environment.

Our results in Table \ref{tab:results_reality} show that the Unicycle-6 agent reached the goal significantly faster than the Point-turn agent on average (48 sec vs. 74 sec), while taking a slightly longer path (7.0 m vs 6.8 m). Additionally, we show that SCT accurately highlights the efficiency of the faster agent, as the Unicycle-6 agent achieves a higher average SCT than the Point-turn agent (57 SCT vs. 37 SCT), while considering SPL alone may make the Unicycle-6 agent seem less efficient than the Point-turn agent (89 SPL vs 91 SPL).

\section{CONCLUSION}

In this work, we propose a new metric, Success by Completion Time. In contrast to other navigation performance metrics like SPL, SCT explicitly considers the dynamics model of the agent's embodiment. This allows SCT to capture how well an agent's learned navigation behavior approximates the ideal behavior afforded by its dynamics, where for SCT, the ideal behavior is defined as one that results in the fastest successful completion times. We also present our RRT*-Unicycle algorithm for approximating the fastest path in an environment with obstacles for an agent with unicycle dynamics, and show that it can be used to evaluate the navigation performance of agents both in simulation and in reality. Using deep reinforcement learning, we train unicycle agents that complete episodes much faster than point-turn agents not by having increased speed, but by learning to leverage their dynamics. We also propose a decaying reward scheme that encourages this behavior and significantly improves SCT performance. Lastly, we show that agents trained solely in simulation with our decaying reward scheme can generalize to reality, with the unicycle agent completing episodes faster than the point-turn agent. 

There are multiple potential future directions of research available to extend our work. Because discrete action spaces are only able to represent a subset of the full space of actions of a dynamics model, it limits the agent's ability to reach the fastest behavior of its dynamics. Thus, research in training unicycle agents with a continuous action space may lead to agents that surpass our agents and achieve near-optimal SCT performance. Another promising direction would be to investigate other models of dynamics, such as car-like robots (that cannot pivot) or legged robots, and use SCT as a better metric for navigation performance than SPL.
\\
\\
\section{Acknowledgements}
\scriptsize{
    The Georgia Tech effort was supported in part by NSF, ONR YIPs, ARO PECASE. The views and conclusions are those of the authors and should not be interpreted as representing the U.S. Government, or any sponsor.
}
\bibliography{references}{}
\bibliographystyle{plain}
\end{document}